%% file: main.tex
\documentclass[sigconf,natbib=false]{acmart}
\usepackage{graphicx}
\usepackage{subcaption}
\usepackage{hyperref}

\usepackage{algorithmic}
\usepackage{graphicx}
\usepackage{gensymb}
\usepackage{textcomp}
\usepackage{xcolor}
\usepackage{stfloats}
\usepackage{array}
\usepackage{arydshln}
\usepackage{tabularx}
\usepackage{multirow}
\usepackage{float}
\usepackage{pgfplots}
\usepackage{mwe}
\newcolumntype{L}[1]{>{\raggedright\let\newline\\\arraybackslash\hspace{0pt}}m{#1}}
\newcolumntype{C}[1]{>{\centering\let\newline\\\arraybackslash\hspace{0pt}}m{#1}}
\newcolumntype{R}[1]{>{\raggedleft\let\newline\\\arraybackslash\hspace{0pt}}m{#1}}

\usepackage{lipsum}
\usepackage{multicol}
\usepackage{graphicx}
\usepackage{pifont}

\usepackage{stackengine}
\usepackage{amsmath}

\def\BibTeX{{\rm B\kern-.05em{\sc i\kern-.025em b}\kern-.08em
    T\kern-.1667em\lower.7ex\hbox{E}\kern-.125emX}}
    
\pgfplotsset{compat=1.17}

\AtBeginDocument{%
  \providecommand\BibTeX{{%
    Bib\TeX}}}

\setcopyright{acmlicensed}
\copyrightyear{2026}
\acmYear{2026}
\acmDOI{}
\acmConference[HRI '26 Workshop: MAgicS-HRI]{Proceedings of ACM/IEEE International Conference on Human-Robot Interaction}{March 16--19, 2026}{Edinburgh, Scotland, UK}
\acmISBN{978-x-xxxx-xxxx-x/YYYY/MM}

\makeatletter
\renewcommand\@copyrightpermission{}
\makeatother


\RequirePackage[
  datamodel=acmdatamodel,
  style=acmnumeric,
  sorting=none
]{biblatex}

\begin{document}

\title{A  Multimodal Framework for Human-Multi-Agent Interaction}

\author{Shaid Hasan}
\authornote{Both authors contributed equally to this research.}
\affiliation{%
  \institution{University of Virginia}
  \city{}
  \country{}}
\email{qmz9mg@virginia.edu}

\author{Breenice Lee}
\authornotemark[1]
\affiliation{%
  \institution{University of Virginia}
  \city{}
  \country{}}
\email{pkb5ne@virginia.edu}

\author{Sujan Sarker}
\affiliation{%
  \institution{University of Virginia}
  \city{}
  \country{}}
\email{zzr2hs@virginia.edu}

\author{Tariq Iqbal}
\affiliation{%
  \institution{University of Virginia}
  \city{}
  \country{}}
\email{tiqbal@virginia.edu}

\renewcommand{\shortauthors}{Hasan et al.}

\begin{abstract}
Human–robot interaction is increasingly moving toward multi-robot, socially grounded environments. Existing systems struggle to integrate multimodal perception, embodied expression, and coordinated decision-making in a unified framework. This limits natural and scalable interaction in shared physical spaces. We address this gap by introducing a multimodal framework for human-multi-agent interaction in which each robot operates as an autonomous cognitive agent with integrated multimodal perception and Large Language Model (LLM)-driven planning grounded in embodiment. At the team level, a centralized coordination mechanism regulates turn-taking and agent participation to prevent overlapping speech and conflicting actions. Implemented on two humanoid robots, our framework enables coherent multi-agent interaction through interaction policies that combine speech, gesture, gaze, and locomotion. Representative interaction runs demonstrate coordinated multimodal reasoning across agents, and grounded embodied responses. Future work will focus on larger-scale user studies and deeper exploration of socially grounded multi-agent interaction dynamics.
\end{abstract}

\maketitle
\input{sections/intro}

\input{sections/system}

\input{sections/result}
\input{sections/discussion}
\input{sections/conclusion}

\printbibliography

\end{document}

%% file: sections/intro.tex
\section{Introduction}
\begin{figure}[!t]
    \centering
    \includegraphics[width=0.45\textwidth]
    {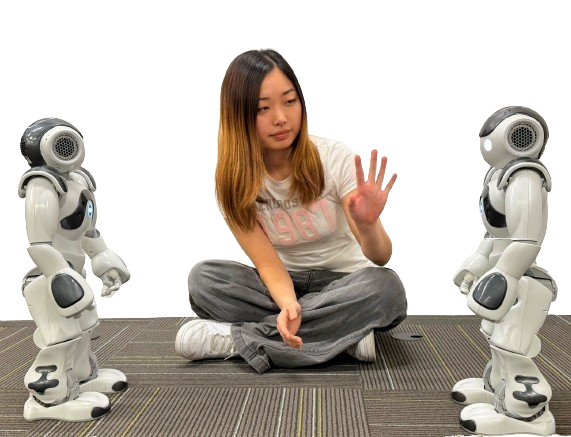}
    \caption{ Multimodal multi-agent human–robot interaction scenario. A single human user engages simultaneously with two humanoid robot agents through speech and non-verbal cues.}
    \vspace{-1.5em}
    \label{fig:interaction_demo}
\end{figure}

Human-robot interaction (HRI) is evolving beyond traditional single-robot, command-based systems toward richer, socially grounded environments in which people interact with multiple autonomous robots through natural communication channels, including speech, gaze, gesture, and embodied motion (Figure \ref{fig:interaction_demo}) \cite{tabatabaei2025gazing, green2025using, zhang2025can, green2025examining}. Multi-agent HRI systems improve scalability, flexibility, and social interaction by allowing groups of robots to work together, divide tasks, and take on different roles \cite{pu2025beatbots, sarker2024cohrt}. This is important in places like public spaces, schools, hospitals, and collaborative workplaces, where people naturally expect robots to act as a team member rather than as independent machines \cite{yasar2022robots,gollob2025envisioning, yasar2024imprint,yasar2024posetron, jiang2023embodied,yasar2023vader,yasar2021improving}.

Effective human-robot interaction fundamentally depends on multimodal communication. As robots become more sophisticated social agents, their ability to perceive and respond through multiple modalities becomes essential for natural, intuitive interaction \cite{islam2025embodied, islam2023patron,islam2023eqa,islam2022caesar, hasan2024m2rl, hasan2024holistic}. In multi-agent settings, multimodality introduces additional complexity and opportunity. Visual feedback and real-time visualization have been shown to enhance human understanding and control in multi-robot systems. However, despite the recognized importance of multimodal interaction, most existing multi-agent HRI systems remain limited in their perceptual and communicative capabilities.

Current multi-agent HRI approaches face three critical limitations in multimodal integration. First, perception is often unimodal or loosely coupled. Many systems rely on either speech commands or visual detection, but rarely integrate these modalities into a unified understanding of the interaction context \cite{gollob2025envisioning, zhang2016optimal, jo2019design}. When multiple modalities are present, they typically operate independently rather than being fused into coherent semantic representations \cite{jo2019design}. Second, embodied multimodal expression remains constrained. While research has explored how robot motion patterns and non-verbal behaviors affect human perception, few systems coordinate speech, gesture, gaze, and locomotion as integrated communicative acts within multi-agent teams \cite{luo2024impact, saadon2025scammed, tan2019one}. Third, multimodal reasoning is disconnected from agent cognition. Existing frameworks often treat perception as a preprocessing step and action as a postprocessing step, with limited integration into the agent's decision-making processes \cite{schombs2025conversation}.

These challenges are made even harder in multi-robot settings. In addition to perception and dialogue, systems must manage team coordination—deciding which robot should respond \cite{zhang2016optimal}, and how actions such as speaking, gesturing, and moving are coordinated without interfering with one another. Each robot also needs to combine multimodal perception with action while sharing the same physical and conversational space. Most existing systems address these issues using simple rules, fixed roles, or symbolic planners, which often lack the flexibility needed to adapt to changing multimodal interaction cues.

To address these challenges, we develop a multimodal multi-agent human–robot interaction system in which each robot operates as an autonomous cognitive agent with integrated multimodal perception and LLM-driven planning grounded in embodiment. Each robot processes speech and visual input using a vision–language model (VLM), combining them into a unified understanding of the interaction context that informs both high-level reasoning and action selection. At the team level, the system employs a centralized coordinator to determine which robot should respond based on multimodal cues and shared context. This enables robots to dynamically take turns, generate coordinated response plans, and execute embodied behaviors, while avoiding overlapping speech or conflicting movements.

%% file: sections/system.tex
\section{System Design}
\begin{figure*}[!t]
\centering
  \includegraphics[width=0.95\textwidth]{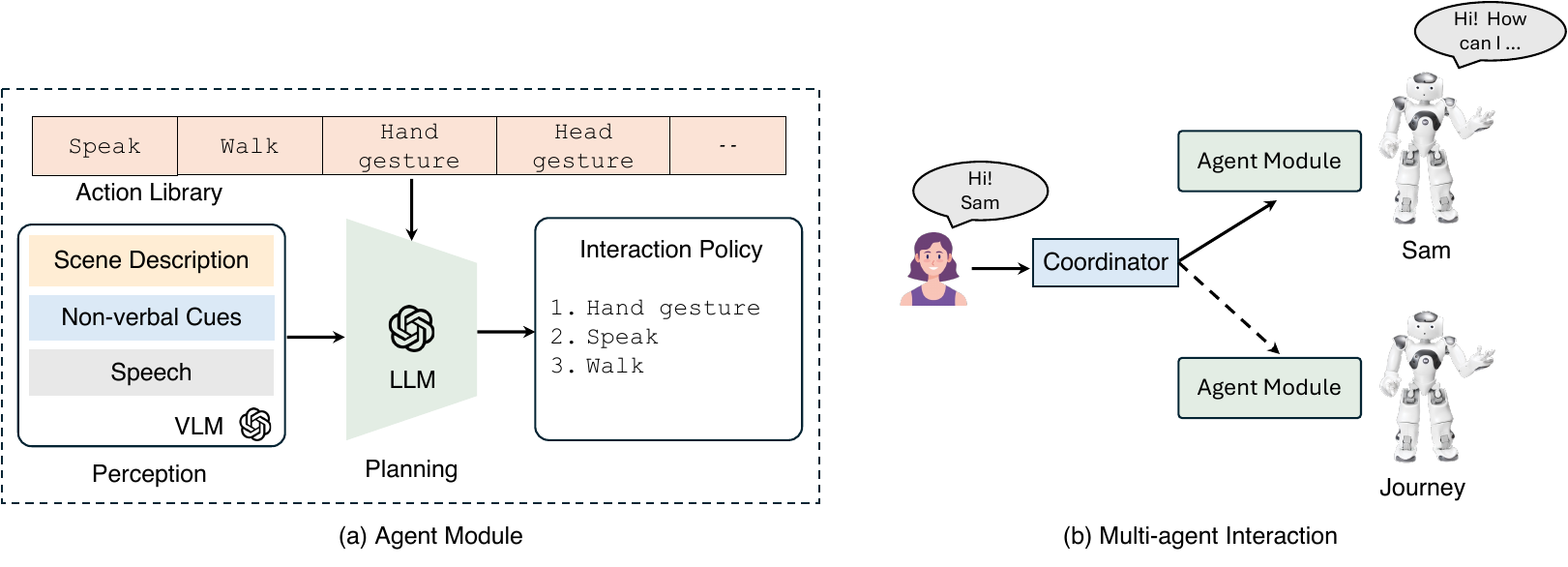}
  \vspace{-1em}
  \caption{ Developed framework overview (a) Internal closed-loop module of a single robot agent, showing multimodal perception, LLM-based planning, and embodied action execution via an action library. (b) Multi-agent interaction setup, where multiple agents (e.g., Sam and Journey) interact with a human user under centralized coordination.}
  \vspace{-1em}
  \label{fig:system}
\end{figure*}

Our developed framework (Figure \ref{fig:system}) is a co-located multimodal multi-agent human–robot interaction system designed to enable natural and coordinated interaction between a single human user and a team of embodied robot agents. In this system, multiple humanoid robots interact with one human in a shared physical space.

\input{sections/system/0_system_intro}

\input{sections/system/1_perception}

\input{sections/system/4_planning}

\input{sections/system/5_action}
\input{sections/system/multi_agent}

%% file: sections/system/0_system_intro.tex
\subsection{Agent Module}

Each robot is modeled as a modular closed-loop cognitive agent composed of three core components: perception, planning, and action execution. These components form a continuous perception–cognition–action loop in which multimodal sensory input is transformed into embodied robot behavior. Perception grounds the agent in the physical and conversational context, planning synthesizes contextual information into structured action policies, and action implements those policies through executable robot behaviors. We describe each module individually below.

%% file: sections/system/1_perception.tex
\subsubsection{Perception}



The Perception Module enables each agent to transform raw multimodal sensory input into a structured semantic observation that serves as the foundation for downstream reasoning and action generation. At each interaction timestep, the agent receives spoken input and visual input captured from its onboard camera. These signals are processed jointly to produce a unified observation representing the current interaction state.

The Perception Module is implemented as a multimodal reasoning component that integrates speech processing, visual sensing, and a vision–language model (VLM). The VLM produces semantically meaningful descriptions of the interaction context, such as grounding spoken references in the physical scene, identifying visible entities, or interpreting user pose. This transforms heterogeneous sensory data into a coherent textual representation that can be directly consumed by language-based reasoning modules.

%% file: sections/system/4_planning.tex
\subsubsection{Planning}
The Planning module serves as the central reasoning component of each agent, integrating perceptual observations, interaction context, and available action affordances to generate embodied response policies. Rather than relying on fixed decision rules or symbolic planners, the system adopts a language-model-driven planning paradigm in which high-level reasoning and action selection are jointly performed by an LLM conditioned on structured cognitive inputs.

At each timestep, the planner receives the current observation, the shared interaction context, and the agent’s action capability library. These inputs are synthesized to produce a policy represented as an ordered list of parameterized actions. Each action corresponds to an executable primitive such as speech, gesture, head movement, or locomotion, along with associated parameters specifying content and execution style. A policy may consist of a single action or a multi-step sequence depending on the interaction context.

A distinctive aspect of the Interaction Planner is that reasoning is constrained by the agent’s embodied action capabilities. Rather than producing free-form text alone, planning is performed with explicit awareness of the available action set, which includes speech, gestures, head movements, locomotion, and vision-based behaviors. The generated policy therefore consists of a structured sequence of action intents, ensuring that every reasoning cycle results in observable robot behavior.

%% file: sections/system/5_action.tex
\subsubsection{Action}
The Action Module implements the agent’s planned behavior in the physical world. Given a parameterized action sequence produced by the planner, the Action Module instantiates and executes each action primitive on the robot, ensuring that every cognitive cycle results in observable behavior. Robot behavior is modeled as compositions of a finite set of reusable action primitives. The action library is designed to cover the core embodied behaviors required for interactive HRI, including verbal communication, nonverbal expression, and movement-based behaviors. In our system, these primitives include speech actions, postural actions (e.g., stand, sit, rest), expressive gestures (e.g., nodding, waving, handshaking), head movement for attention and camera framing, locomotion commands, and simple arm or hand motions such as pointing and hand open/close. Each primitive accepts parameters that specify the content and style of execution, such as utterance text and volume for speech, gesture type and speed, head pan or tilt angles, or walking direction and magnitude.

The Action Module converts each action–parameter pair into a concrete robot command and dispatches it to the execution layer. Because actions are parameterized, the module acts as an interpreter that validates parameters, ensures compatibility with the robot’s capabilities, and executes actions with appropriate timing and sequencing constraints. Multi-step policies are handled by executing actions in order, allowing the planner to express compound behaviors such as “gesture → speak → nod,” or to interleave perception-driven actions with communication. Execution returns a status signal indicating success or failure for each action, supporting robustness in interactive settings and enabling graceful degradation. In addition, certain actions directly influence subsequent perception—for example, head movement affects camera viewpoint—thereby closing the sensorimotor loop.

%% file: sections/system/multi_agent.tex
\subsection{Multi-Agent Interaction}
Our framework adopts a centralized coordination mechanism to regulate interaction among multiple agents while preserving decentralized cognition. Each agent independently maintains its own perception–planning–action loop as described in Section 2.1, while a shared coordinator implements conversational turn-taking and agent participation.

At each interaction timestep, every agent produces its local observation through its Perception Module, and the system maintains a global interaction context representing conversational history. The centralized coordinator evaluates how appropriate it is for each agent to respond given its current observation and the shared context. This evaluation is implemented using a language model that produces response likelihood scores for all agents. Agents whose scores exceed a predefined threshold are selected to participate. If a single agent is selected, it responds alone; if multiple agents are selected, they respond sequentially. Each selected agent independently generates its own embodied interaction policy using its planner, which is then executed through its Action Module.

Although response likelihood could in principle be computed in a decentralized manner, centralized coordination is used to enforce global interaction constraints essential for embodied multi-robot settings. Central arbitration prevents simultaneous speech and conflicting physical actions, ensures deterministic turn ordering, and resolves directed addressing when a user explicitly names a robot. Achieving these guarantees in a fully decentralized system would require complex inter-agent consensus mechanisms.

%% file: sections/result.tex
\section{System Demonstration and Results}
\begin{figure}[!t]
    \centering
    \includegraphics[width=0.48\textwidth]
    {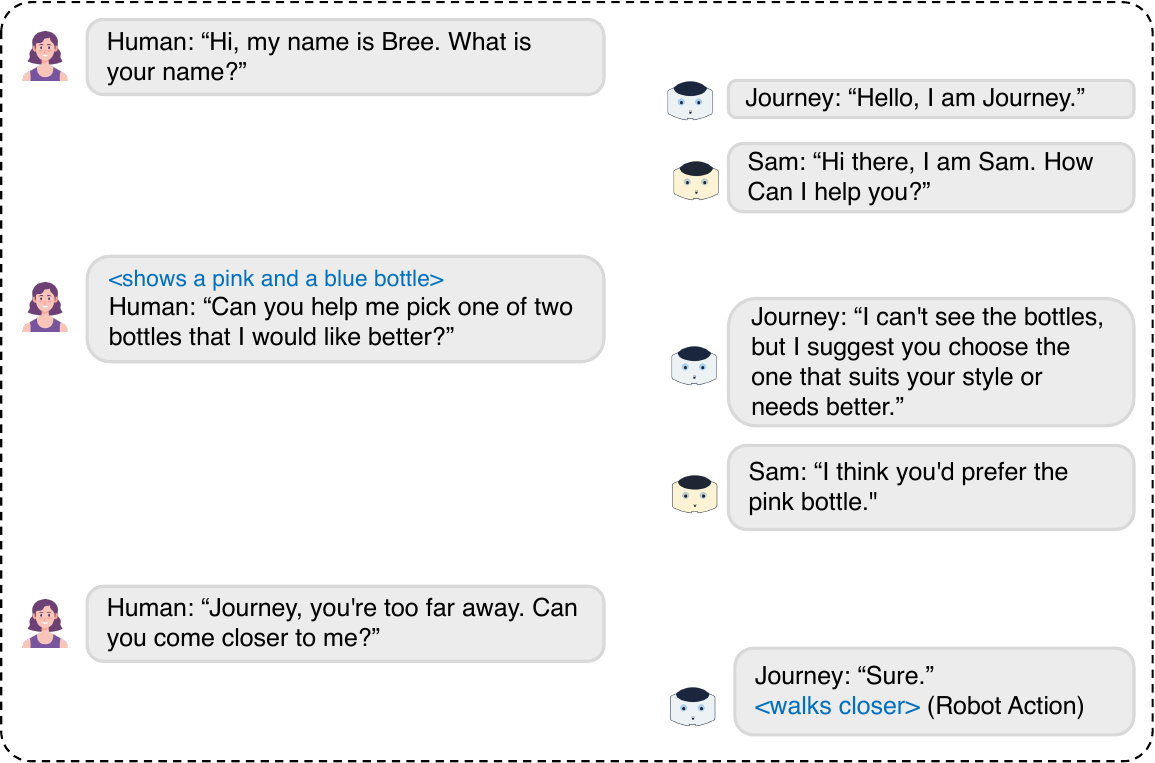}
    \caption{ A representative interaction run from our system demonstration. A human user engages two humanoid robots in shared space through speech and visual cues. The central coordinator selects the most contextually relevant agent(s) to respond, enabling sequential, non-overlapping dialogue and grounded physical actions such as verbal replies and locomotion.}
    \vspace{-1.5em}
    \label{fig:interaction}
\end{figure}

Our proposed framework shows effective integration of multimodal perception, and coordinated multi-agent response. From the interaction example shown in Figure~\ref{fig:interaction}, we observe that the system successfully processes spoken dialogue, non-verbal cues, and directed addressing within a shared physical environment.

At the start of the interaction, the human introduces herself, and both robots respond sequentially, demonstrating coordinated turn-taking without overlapping speech. This indicates that the centralized coordinator is able to regulate participation among agents while preserving individual responses. When the user asks for help choosing between two bottles, the robots generate distinct but contextually grounded replies: \textit{Journey} acknowledges its visual limitation and provides a general recommendation, while \textit{Sam} offers a personalized suggestion. This illustrates distributed reasoning across agents, where each robot interprets the same interaction through its own perceptual and contextual understanding.

Finally, when the user explicitly addresses \textit{Journey} and requests it to come closer, the system correctly resolves directed speech, and \textit{Journey} executes the physical action while confirming verbally. This behavior demonstrates successful grounding of language into embodied action. Overall, the interaction shows that our developed framework supports multimodal understanding, coordinated multi-agent participation, and execution of physical behaviors in human-robot interaction scenarios.

%% file: sections/discussion.tex
\section{Discussion}

Our framework explores how multimodal perception, embodied action, and centralized coordination can be integrated to support human–multi-agent interaction. Through multiple demonstration runs, we observed interaction patterns that suggest several design insights. Our demonstrations suggest that multimodal perception strongly influenced how interactions unfolded between human and robot teams. Although speech and vision were combined into a single representation, real-world factors, such as occlusions, lighting changes, and speech recognition noise introduced ambiguities during interaction. In practice, these ambiguities appeared as misunderstandings rather than clear sensing errors. In multi-agent settings, they also affected which robot chose to respond, indicating that perception quality can shape how agency and responsibility emerge within the team.

Delays from the LLM and VLM, along with the robot’s limited physical expressiveness, clearly influenced how interactions unfolded. Slow responses and constrained movements affected how attentiveness and competence were conveyed. Rather than treating latency and embodiment as purely technical issues, we view them as communicative signals that shape turn-taking and engagement. Providing incremental feedback and better aligning action policies with the robot’s expressive capabilities may help maintain conversational flow and make robot intentions clearer.

Across our demonstrations, interactions reflected the system operating as a coordinated team rather than as isolated agents. Although centralized arbitration prevented overlapping speech and conflicting actions, effective human–multi-agent communication depended on whether this coordination was visible externally. Signs of mutual awareness and shared context appeared to influence how collective behavior was understood.

While our current implementation demonstrates coordinated interaction between two humanoid agents, scaling the framework to larger robot teams can introduces computational challenges. The centralized coordination mechanism evaluates response likelihoods for each agent at every timestep, and as the number of agents increases, arbitration complexity and latency may grow, potentially affecting conversational fluency and action.


%% file: sections/conclusion.tex
\section{Conclusion}
We presented a multimodal multi-agent human–robot interaction system that integrates vision–language perception, LLM-driven planning, and embodied action within a centralized coordination framework. By modeling each robot as an autonomous cognitive agent while regulating turn-taking at the team level, the system enables coherent, socially grounded interaction between a human user and multiple robots in shared physical spaces. Representative interaction runs demonstrate multimodal understanding, coordinated agent participation, and grounded embodied responses. This work provides a practical foundation for future multi-robot systems, and we plan to extend it through larger user studies and deeper exploration of long-horizon, socially grounded interaction.